\pdfoutput=1

\documentclass[11pt]{article}

\usepackage{acl} 

\usepackage{times}
\usepackage{latexsym}

\usepackage[algoruled,ruled,vlined,noend]{algorithm2e}       
\usepackage{times}
\usepackage{latexsym}
\usepackage{pgfplots}
\usepackage{graphicx}
\usepackage{enumitem}
\usepackage{xcolor,colortbl}  
\usepackage{tikz}
\usepackage{diagbox}
\usepackage{tkz-tab}
\usepackage{caption}
\usepackage{booktabs} 
\usepackage{latexsym}
\usepackage{amssymb}
\usepackage{amsmath}
\usepackage{subcaption}
\usepackage{natbib}
\usepackage{blindtext}
\usepackage{url}
\usepackage{color}

\usepackage{url}
\usepackage{hyperref}    
\usepackage{cleveref}
\SetAlFnt{\small}
\SetAlCapFnt{\small}
\SetAlCapNameFnt{\small}
\usepackage{ntheorem}

\usepackage{booktabs, tabularx}
\usepackage{tfrupee}  

\newcommand{\checklist}{\textsc{CheckList}\xspace}

\usepackage[colorinlistoftodos]{todonotes}

\usepackage[T1]{fontenc}

\usepackage[utf8]{inputenc}

\usepackage{microtype}

\title{SALTED: A Framework for SAlient Long-Tail Translation Error Detection}

\author{Vikas Raunak\qquad Matt Post\qquad Arul Menezes\\
~Microsoft \\
\texttt{\{viraunak, mattpost, arulm\}@microsoft.com}}

\begin{document}
\maketitle  
\begin{abstract}

Traditional machine translation (MT) metrics provide an average measure of translation quality that is insensitive to the long tail of behavioral problems in MT.
Examples include translation of numbers, physical units, dropped content and hallucinations. 
These errors, which occur rarely and unpredictably in Neural Machine Translation (NMT), greatly undermine the reliability of state-of-the-art MT systems.
Consequently, it is important to have visibility into these problems during model development.
Towards this direction, we introduce SALTED, a specifications-based framework for behavioral testing of MT models that provides fine-grained views of salient long-tail errors, permitting trustworthy visibility into previously invisible problems.
At the core of our approach is the development of high-precision detectors that flag errors (or alternatively, verify output correctness) between a source sentence and a system output.
We demonstrate that such detectors could be used not just to identify salient long-tail errors in MT systems, but also for higher-recall filtering of the training data, fixing targeted errors with model fine-tuning in NMT and generating novel data for metamorphic testing to elicit further bugs in models.


\end{abstract}

\section{Introduction}
\label{sec:sec1}

The development of Machine Translation (MT) systems is typically guided by performance metrics \cite{bleu, comet} computed on small, curated test sets, wherein system quality is often reduced to a single number. 
While metrics are useful for characterizing the average performance of a system, they do not provide fine-grained visibility into rarer error categories. As a result, researchers do not have a reliable way to gauge whether and to what extent a system may exhibit a wide range of negative behaviors, such as hallucinations, dropped content, or the sporadic mistranslation of important content such as names or physical units.
However, such salient `long-tail' errors undermine the reliability of MT systems, and are increasingly important to curtail in an era where MT output is often indistinguishable from that of humans \cite{martindale}.

A second related problem is that many of these behaviors are rare enough that they will not be observed on standard test sets (which typically only number a few thousand sentence-pairs), even if reliable detection via references is feasible.
Owing to this rarity, the detection methods that require source-reference pairs are not useful, besides being non-scalable.
Consequently, a real gap in machine translation evaluation is having fine-grained analysis that can scale to large, unannotated datasets. In this work, we propose SALTED as a framework to tackle these challenges. Our main contributions are as follows:

\begin{enumerate}
    \item We explore Behavioral Testing \cite{beizer} as a means to provide fine-grained measurements of salient long-tailed errors in MT, while addressing the challenges of rarity and scalability in obtaining those measurements.
    \item We propose an iterative, specifications-based process for obtaining reliable measurements through high-precision detectors and demonstrate their utility on seven MT error classes across both research and commercial systems.
    \item We demonstrate that detectors are amenable to multiple applications in MT, including higher-recall training data filtering, system-comparisons, metamorphic testing and fixing errors through fine-tuning on synthetic data.
\end{enumerate}

\begin{table*}[ht]
\begin{tabularx}{\linewidth}{ l X c }
\toprule
\textbf{Property} & \textbf{Correct Behavior Specification for Translation} &  \textbf{Violation Example}    \\
\midrule
Physical units & The model should translate the exact unit in the target language (abbreviations are allowed). &  yards $\to$ Meters \\
Currencies & The model should translate the exact currency in the target language (both symbol abbreviations and expansions are allowed). &  USD $\to$ € \\
Large Numbers & Large Numbers in text form should appear in the same denominations in the output. & trillions $\to$ millions
\\ 
Web Terms & URLs and Web Addresses should be copied as is from the source to target, without any translation. & www.bbc.en $\to$ www.bbc.de      \\
Numerical values & The number in a numerical value should not change beyond an allowed set of transformations (e.g., time format change, change of separators, decimal point change, number system change etc.). &  24.70 $\to$ 2,470 \\
\midrule
Coverage & The model should translate the entire semantic content of the source sentence. &  My friend Bob $\to$ Mi amigo \\
Hallucinations & The model should not produce translated content that is not grounded in the source sentence. &  Hello $\to$ Hola ha ha ha ha \\
\bottomrule
\end{tabularx}
\caption{\textbf{Behavior specification}: The first step in building a detector is to specify the correct expected behavior.}
\label{tab:spec_table}
\vspace{-1.0em}
\end{table*}

\section{The SALTED Approach}
\label{sec:sec2}

Behavioral Testing \cite{beizer} concerns itself with testing the input-output behavior of systems, without leveraging any knowledge about the system's internal structure. For behavioral testing of Natural Language Processing (NLP) models, \checklist \cite{checklist} proposes a process to construct test cases for evaluating different linguistic capabilities, each test case being a specified input with the associated ground truth label(s). This approach does not generalize to NMT for several reasons; (a) there could be multiple valid translations of the same input (b) errors are highly contextual: the same phenomenon may be translated accurately in one sentence, but inaccurately in another (c) errors are unstable: different model iterations may manifest errors in different sentences (d) errors are rare: a particular  mistranslation may manifest itself only once in a million sentences. For these reasons, an annotated set of test cases is not just challenging to construct, but also instantly obsolete. In SALTED, we propose a different approach to behavioral testing for NMT. Instead of relying on test cases, we translate millions of sentences and apply \emph{detectors}. Each detector is written to detect a specific class of error, based on a specification of correct behavior. 

More concretely, \textit{a detector is an algorithm, which given an input-output sentence pair returns a boolean value indicating the presence of an error condition with very \textbf{high precision}.} The proposed detectors lie at the extreme non-trivial end of the precision-recall curve and emphasize very high precision in order to make the ensuing measurements trustworthy. At a high-level, we construct detectors by iteratively narrowing the error specifications until absolute precision is achieved on a large development set. While this means that a number of \textit{potentially} erroneous output instances would not be flagged, we gain the advantage that the resulting detector could now act as a trustworthy measurement of a specific error category, a key property of useful measurements \cite{hand2016measurement}.

\section{Behavior Specification for MT}
\label{sec:sec3}
If we cannot specify the correct behavior, we cannot verify output correctness \cite{formal_spec}. Therefore, the first step of constructing a detector is to specify the desired behavior that the model must satisfy with respect to the translation of certain salient content or property. Table \ref{tab:spec_table} lists the behavior specifications for all the detectors we implement. Note that the choice of salient long-tailed error classes itself depends on the system developer. We determined the 7 error classes in Table \ref{tab:spec_table} owing to their disproportionate impact on user trust, since each of these specifications if violated could lead to serious consequences for user consumption of translations. The desired behavior for some cases (e.g., URL translation) is unambiguous. However, in the case of expressions such as physical units or currencies, an unambiguous specification is not readily apparent, e.g., NMT models are quite capable of learning to translate `10 miles' to `16 km' from parallel data. While this may be desirable for localization purposes, and successful in common cases, it could lead to dangerous inaccuracies in rarer cases, since NMT models shouldn't be trusted to do mathematics consistently and correctly, even if the conversion rules do not vary across time. Therefore, in general, our behaviour specifications require that such expressions be left semantically unchanged. 
This however applies only to the preservation of semantic content; as evident from Table \ref{tab:spec_table}, the specification allows changes to digit separators and to number systems (such as those between English and Chinese) in the case of numerical values. Further, as we show in section \ref{sec:designing_detectors_main} this step of behavior specification is itself subject to iteration within the process of building detectors. However, an explicit enumeration of the boundary between desired and undesired behavior across different salient content types or properties provides us a starting point for a detector implementation.

\section{Designing Detector Algorithms}
\label{sec:designing_detectors_main}

Building detectors from specifications is not always an easy process.
A first implementation for most content types inevitably yields large numbers of false positives. It is here that our focus on \emph{precision} over recall serves as a useful guide. In addition to producing detectors whose results are more trustworthy, it simplifies the task, since we can narrow in on a subset of settings that could be detected with a high certainty of correctness. Here we illustrate this process, concurrently with an example.

\subsection{Example: Physical units translation}

We consider the identification of errors in the translation of physical units, such as meters, feet, etc. for English $\rightarrow$ German translations. We decompose the process of constructing the detector for physical units translation errors into three steps: behavior specification, resource construction and checking for specified behavior. Each of these steps is iterated upon by quantifying the precision of error detection through human evaluation. The development iterations are done on a large initial set of monolingual source sentences and their translations generated by a MT system and the development is halted until absolute precision is achieved on this corpus. Finally, a `test' phase human evaluation is conducted by \textit{varying both} the monolingual data as well as the MT system, to ascertain the final precision of the developed detectors.



\paragraph{Behavior Specification} In this case, we start with the specification in Table \ref{tab:spec_table}, i.e., the desired behavior is that the physical unit measurement in the source be `carried through' without changes in the target language. For example, `10 feet' getting translated to `10 meters' or `10 miles' getting translated to `16 km' are both errors.



\begin{table}[ht!]
    \begin{tabularx}{\linewidth}{l r}
        \toprule
\textbf{Token Transformation Table Entry} &  \textbf{Type}    \\
        \midrule
meter $\rightarrow$ meter, m  & dist    \\
 mm $\rightarrow$ Millimeter, Millimetern, mm & dist  \\
 feet $\rightarrow$ Fuß, Füße, Fußende & dist  \\
  mile $\rightarrow$ meile, meilen & dist  \\
  km² $\rightarrow$ km², Quadratkilometer & area  \\
  sq.ft. $\rightarrow$ sq.ft., Quadratfuß, Quadratfuße & area  \\
        \bottomrule
    \end{tabularx}
    \caption{A partial view of the \textbf{Token Transformation Table} constructed for use in physical unit detector. Each row comprises of allowed token transformations, along with a token `type' annotation (used in section \ref{sec:sec7}).}
    \label{tab:transformation_table}
    \vspace{-1.0em}
\end{table}

\paragraph{Transformation Table} Once the desired behavior for the detector has been specified, the next step is to build the relevant resources in order to facilitate checking for the desired behavior on an arbitrary sentence pair. Table \ref{tab:transformation_table} illustrates the resource constructed in this case: a `Transformation Table' of relevant source tokens which maps a source token to its set of potential translations (transformations). As we will demonstrate later, building this `Transformation Table' is quite tractable for expressing tests that require checking for token-level transformations in translations. Further, during the process of construction of the `Transformation Table', we also \textit{annotate the type of the source token} (we explain its utility in section \ref{sec:sec7}). 

\paragraph{Checking for Specified Behavior} Once the `Transformation Table' has been constructed, the detector checks for the desired behavior as follows: if a source token in the transformation table is found in an input sequence, then the output sequence must contain one of the possible mappings of the source token. We also enforce that the source token must be delimited by space and that the potential target tokens need not be delimited by space in the output sequence. Note that by selectively relaxing the check on the target side, we protect against lowering detector precision due to non-semantic/formatting changes i.e., we allow transformations such as `10 km' $\rightarrow$ `10km'. 


\begin{table}
  \centering
\setlength\tabcolsep{1.00pt}
  \begin{tabular}{lcr}
    \toprule
    \textbf{Iteration}  & \textbf{Algorithmic Changes} & \textbf{Precision} \\
    \midrule
    1 & None, Initial Conditions & 72.0  \\
    2 & Numeric Measurements Only & 94.0  \\
    3 & Fixes in Transformation Table & 100.0 \\ \midrule
  \end{tabular}
   \vspace{-0.75em}
  \caption{\textbf{Iteration vs Precision} on Physical Units Detector, measured using Human Evaluation on 100 cases flagged by error by the detector}
  \label{tab:iteration}
  \vspace{-1.0em}
\end{table}

\paragraph{Iterations vs Precision} At each iteration, we apply the resulting detector on translations of a 1M random sample of the WMT20 English monolingual data (dataset/system details are presented in section \ref{sec:dataset_details}) and measure precision by conducting a human evaluation of 100 flagged cases, selected at random at each iteration. Table \ref{tab:iteration} shows the resulting precision of the physical units detector across 3 iterations. At iteration 1, we observed a number of false positives pertaining to idiomatic expressions ("missed by a mile") and approximations ("a few yards further"), which were adequately translated despite missing the exact unit translation. Therefore, going from iteration 1 to 2, we narrowed our error detection only to the cases when a physical unit was preceded by a number (either in text or numeric form). This helped us avoid false positives due to the alternate senses and idiomatic uses of certain units such as feet, leading to significantly higher precision in iteration 2. Going from iteration 2 to 3, guided again by the goal of improving precision, we added/fixed entries in the transformation table to avoid false positives. The false positives obtained during each iteration of the detector are presented in appendix \ref{sec:appendix_false_positives}. We halted the manual iterations upon achieving 100\% precision in human evaluation. Table \ref{tab:sota_examples} presents an example of physical unit error flagged using the resulting detector. Further, we present the results of the final test evaluation in appendix \ref{sec:appendix_test_detectors}.

\subsection{Full Suite of Detectors}
\label{sec:sec4.2}

\begin{table*}[ht]
    \begin{tabularx}{\linewidth}{l X}
        \toprule
    \textbf{Detector} & \textbf{Source-Translation Instance}  \\ \midrule
    
Physical Unit & Teacher's hallway song and dance reminds students to stay \colorbox{yellow}{6 feet} apart. \\
& Lehrer Flur Lied und Tanz erinnert die Schüler zu bleiben \colorbox{orange}{6 Meter} auseinander. \\ \hline

Currency & Floorpops Medina Self Adhesive Floor Tiles,  \colorbox{yellow}{£14} from Dunelm - buy now \\
 &  Floorpops Medina selbstklebende Bodenfliesen, \colorbox{orange}{15 €} von Dunelm günstig kaufen \\ \hline  


Numerical Value & Kerridge has been an outspoken defender of his industry \colorbox{yellow}{throughout 2020}, but it was an angry Instagram post that may have made the most difference.\\
& Kerridge war das ganze \colorbox{orange}{Jahr über ein} ausgesprochener Verteidiger seiner Branche, aber es war ein wütender Instagram-Post, der möglicherweise den größten Unterschied gemacht hat.\\ \hline  

Coverage & Ben Cooper QC suggested it was unfair that the conspiracy theorist was arrested on \newline \colorbox{yellow}{May 30 while no arrests were made for breaches of lockdown restrictions at a} \newline
\colorbox{yellow}{Black Lives Matter protest taking place on the same day}. \\
 &  Ben Cooper QC hielt es für unfair, dass der Verschwörungstheoretiker am 30. \\ \hline

Hallucination & \colorbox{yellow}{The Cougars are supposed to play No.} \\
 & \colorbox{orange}{== Weblinks ==== Einzelnachweise ==} \\ 

\bottomrule
    \end{tabularx}
    \caption{\textbf{Detector Output examples} from the 100K WMT20 Monolingual-Evaluation set: All rows show errors made by commercial systems, as flagged by various detectors. The last row shows an error by the Microsoft system, rest show errors made by the Google system. All public APIs were accessed on January 10, 2021.}
    \label{tab:sota_examples}
    \vspace{-1.0em}
\end{table*}

The space of detector algorithms is not at all constrained by how they function as long as the contract of high-precision is satisfied. However, in this work, we mainly consider two kinds of detectors, namely token-level and sequence-level detectors.

\paragraph{Token-level Detectors} Token-level detectors represent a generalization of the detector instance described in section \ref{sec:designing_detectors_main}. Token-level detectors rely on language-pair specific transformation tables and as such, are well suited for testing the transformations of source tokens pertaining to a number of salient content types. Following the same methodology for constructing the \textbf{Physical Units} detector from section \ref{sec:designing_detectors_main}, we construct detectors for evaluating the translation of salient tokens corresponding to three more content types in Table \ref{tab:spec_table}, namely \textbf{Currencies} (e.g., USD, \$),\textbf{ Large Numbers} (e.g., millions, billions) and \textbf{Web Terms} (e.g., URLs and web address terms such as https, www). Additional implementation-level details regarding the token-level detectors are provided in appendix \ref{sec:appendix_token_detectors}. 

We also construct a token-level detector to test the translation of \textbf{numerical values}. Here, instead of a fixed transformation table, the transformation table is generated on the fly per instance. For this numerical values detector, we extract contiguous numerical values (digits) from the input sequence, condense the value's representation into a single token (by removing separators) and allow for a range of possible transformations of the numerical value, which are then checked against the output. The primary transformations considered are time conversions and date conversions. The inherent logic in this case is the same as for previous token-level detectors, except that instead of a transformation table, we construct transformation functions which are applied on the fly to generate the table. This approach of behavioral testing the translation of numerical values is quite general, unlike the explicit construction of test cases in \cite{numerical}.

\paragraph{Sequence-level Detectors} Sequence-level detectors do not rely on explicitly constructing language-pair specific transformation tables/functions and instead leverage more general mechanisms or resources. Such detectors are best suited for properties wherein the correct behavior can be verified using the artifacts computed from the input and the output sequences. We construct detectors for two sequence-level properties/phenomena: namely, coverage and hallucinations. 

For building the \textbf{coverage detector}, we measure the number of content words (non-stopwords, non-punctuations) left unaligned using Awesome-Aligner \cite{awesome_aligner} with multilingual BERT as the contextual embedding model. An input-output pair is flagged as an error when the number of unaligned content words exceeds a threshold. This threshold for unaligned content words is set in proportion to input sentence length.

For constructing the \textbf{hallucination detector}, we start from the quantitative definition of hallucinations from \citet{curious} and adjust the thresholds for target-repeat and oscillatory hallucination detectors until high precision is achieved. 

Appendix \ref{sec:appendix_test_detectors} provides the results of the final `test' evaluation for each of the detectors, while appendix \ref{sec:appendix_detector_suite} provides additional detector details.

\section{Evaluations using SALTED}
\label{sec:sec5}

Having constructed seven high-precision detectors, we now wish to apply them to commercial systems to see whether we can discover any problems. We take a sample of 100K sentences from a larger 1M monolingual corpus (detailed in section \ref{sec:dataset_details}) and translate them with Google, Microsoft, and Amazon’s systems by way of their paid public APIs. Table \ref{tab:sota_statitiscs} shows the raw counts of erroneous translations while Table \ref{tab:sota_examples} presents some instances of the flagged errors from different detectors. 

What these results show is that the long-tailed errors are quite pervasive across NMT systems, despite being very rare (only \textbf{0.3\% incidence rate} for Google, based on Table \ref{tab:detector_errors}). To further validate this inexpensively with more data, we translate the full 1M monolingual corpus using the WMT21 News translation task winning system, the results (both raw counts and examples) of which are presented in appendix \ref{sec:appendix_wmt21_examples}. 

\begin{table}[ht!]
    \centering
    \setlength\tabcolsep{1.5pt}
    \begin{tabular}{lrrr}
    \toprule 
    Property & GOOG & MSFT & AMZN \\
    \midrule
    Coverage & 165 & 1 & 8 \\
    Hallucinations & 0  & 5 & 0 \\
    Physical Units & 46 & 6 & 15 \\
    Currencies & 4 & 1 & 0 \\ 
    Large Numbers & 7 & 1 & 4 \\ 
    Web Content & 0 & 0 & 0 \\ 
    Numerical Values & 96 & 11 & 27 \\ \midrule
    Total Errors & 318 & 25 & 54 \\
    \bottomrule
    \end{tabular}
    \caption{Counts of \textbf{Erroneous Translations} found by Detectors in the 100K WMT20 Monolingual Eval Set.}
    \label{tab:sota_statitiscs}
    \vspace{-1.0em}
\end{table}

\section{System Comparisons \& Data Filtering}
\label{sec:sec6}

A general trend in NMT is the susceptibility of trained systems to even small amounts of noisy data \cite{ott-etal-2019-analyzing}. 
We investigate whether detectors--optimized for error precision, rather than recall--can work as effective filters to improve systems.

\subsection{Datasets and Systems}
\label{sec:dataset_details}

\paragraph{Training and Evaluation Datasets} We conduct experiments on the WMT20 News Translation (English-German) task benchmark \cite{wmt-2020-findings}. The standard WMT20 test set is used for measuring general translation performance. For behavioral testing at scale using detectors, we create a Monolingual-Evaluation set of 1M English sentences randomly sampled from the WMT20 monolingual data. Due to cost constraints (e.g., in evaluating public NMT systems), we also sampled a smaller 100K Monolingual-Evaluation set. Model and training details are presented in appendix \ref{sec:appendix_section6_details}.

\paragraph{Systems} We trained three systems (Table \ref{tab:filtering}) each with a different training data-filtering algorithm:
\begin{itemize}
\item \textbf{Unfiltered (UN-F)}: The full English-German parallel training dataset provided by the WMT20 benchmark is used for training.
\item \textbf{Standard (STD-F)}: We replicate the bitext filtering pipeline of \citet{tencent}, one of the top WMT20 systems. Herein, sentence-pair filtering based on maximum allowable sentence-length ratio (1:1.3) and reverse sentence-length ratio (1.3:1) is applied on the unfiltered corpus, alongside filtering sentences greater than a maximum word length (150). A language-id filter \cite{fastext-langid} is also applied, which checks if the source and target languages are in the correct languages.
\item \textbf{Detector-Based (DB-F)}: For this system, filtering as per \citet{tencent} is replaced by filtering using the full suite of detectors; except for the use of language-id which is the same as in \citet{tencent}.
\end{itemize}

\paragraph{System Comparisons} The comparison of the Unfiltered (UN-F) and Standard (STD-F) systems in Table \ref{tab:filtering} shows that the unfiltered system gets higher BLEU and lower TER on the WMT20 test set, apparently indicating that filtering didn't have any benefits. However, when the full suite of detectors is run on the 1M Monolingual-Evaluation set outputs, the impact of filtering becomes apparent. The Standard system incurs significantly fewer coverage errors and hallucinations as well as fewer errors in the translation of numerical values and currencies. These measurements bring to light the previously hidden impact of filtering since the standard metrics aren't able to capture these trade-offs in model behaviors, achieving only similar scores.

\begin{table}
  \label{tab:table3}
  \centering
\setlength\tabcolsep{4.0pt}
  \begin{tabular}{lrrr}
    \toprule
    \textbf{Measurement} & \textbf{UN-F} &\textbf{STD-F} &\textbf{DB-F} \\
    \midrule
    Training Data &  \textbf{48.2M}  & 36.9M & 41.7M  \\ \midrule
    BLEU $\uparrow$ &     32.4    & 31.4 & \textbf{32.9}    \\
    ChrF2++ $\uparrow$ &     58.4    & 58.0 & \textbf{58.8}    \\
    COMET $\uparrow$ &  42.05  & 38.12 & \textbf{45.79}   \\ 
        TER $\downarrow$ &  54.5  & 55.5 & \textbf{54.2}   \\ \midrule
        Coverage $\downarrow$ & 742 & \textbf{309} & 365  \\
    Hallucinations $\downarrow$ & 37 & \textbf{0} & 8 \\
    
      \midrule
    Physical Units $\downarrow$ & 141 & 151 & \textbf{126}  \\
    Currencies $\downarrow$ & 17 & \textbf{7} & 13  \\
    Large Numbers $\downarrow$  & 113 & \textbf{60} & 67  \\
    Web Terms $\downarrow$  & 43 & 39 & \textbf{33}  \\ 
    Numerical Values $\downarrow$  & 1,000 & 503 & \textbf{429}  \\ \midrule
  \end{tabular}
  \caption{\textbf{Metric Based} System Comparisons on the WMT20 Test set and \textbf{Detector Based} system comparisons on the 1M Mono-Eval Set for the three systems. Note that $\downarrow$ implies lower is better, $\uparrow$ implies otherwise.}
  \label{tab:filtering}
  \vspace{-1.0em}
\end{table}

\begin{table*}[ht]
    \begin{tabularx}{\linewidth}{ l X l}
        \toprule
    \textbf{Sequence Type} & \textbf{Instance} & \textbf{Algorithm Step}  \\
        \midrule
Source &  The plesiosaur teeth it self is about 43 \textbf{mm} long. &  \\
Reference &  Der Plesiosaurier Zahn selber misst etwa 43 \textbf{mm}. &  Sentence Selection \\
\hline
Templatized Source &  The plesiosaur teeth it self is about 43 [\textit{VAL}] long. &  \\
Templatized Reference &  Der Plesiosaurier Zahn selber misst etwa 43 [\textit{VAL}]. & Templatization \\ \hline
Meta Source Instance &  The plesiosaur teeth it self is about 43 \textbf{\textit{feet}} long. & \\
Meta Reference Instance &  Der Plesiosaurier Zahn selber misst etwa 43 \textbf{\textit{Fuß}}. & Type Substitutions\\
        \bottomrule
    \end{tabularx}
    \caption{\textbf{Meta-Corpus} instance generation example using the physical units detector in Algorithm \ref{algo:meta_corpus}.}
    \label{tab:meta_corpus}
    \vspace{-1.0em}
\end{table*}

\paragraph{Data Filtering using Detectors}

The third column in Table \ref{tab:filtering} shows the measurements for the system trained on data filtered using the suite of detectors. The results show that the DB-F system achieves higher BLEU than both the Standard and Unfiltered systems, while yielding similar results to STD-F on the various long tail error categories. This indicates the benefits of a more targeted approach to filtering through high-precision detectors, which helps the model strike a better trade-off between preserving general model performance and preventing fine-grained translation errors. However, it is also clear that filtering alone is not sufficient in reducing salient long-tail errors, e.g., the number of errors in the physical unit category is relatively unchanged. In section \ref{sec:sec8}, we show how SALTED could be used to inject correct model behavior through data synthesis.

\subsection{Discussion}
The results show that a very targeted multi-dimensional view of model behavior could be developed through the use of detectors, bringing visibility into fine-grained model performance not evident through traditional metrics. The results also show that the same detectors could be used as an alternative to standard corpus filtering \cite{tencent}, leading to better model performance. 

\section{Metamorphic Testing}
\label{sec:sec7}

The low incidence rates of long-tail error necessitate large amounts of data in order to elicit errors.
The SALTED framework further helps address this problem via \emph{metamorphic tests}, which produce new test inputs by modifying an input instance in systematic ways, leading to high-density error discovery.

\paragraph{Experiment} Given a token-level detector and an initial corpus of monolingual (source) sentences, if a source token in the detector's transformation table is found in a source sentence, delimited by space on either sides, we create new instances by substituting that token with others of the same type (as annotated in Table~\ref{tab:transformation_table}).
For example, a sentence with the word `meters' can be changed to one with the word `yards'.
The new sentences can then be translated by a system, and the detectors applied to these novel (input,~translation) pairs.

\begin{table}[ht!]
    \centering
    \setlength\tabcolsep{4.5pt}
    \begin{tabular}{lrr}
    \toprule 
    Property & New Sentences & Novel Cases\\ \midrule
    Physical units & 204,029 & 4,503 \\   
    Currencies & 1,232,988 & 775 \\  
    Large Numbers & 7,885 & 42 \\   
    Web Content & 8,238 & 196 \\  
     \bottomrule
    \end{tabular}
    \caption{\textbf{Metamorphic Testing}: New instances produce novel errors in our research system.}
    \label{tab:meta_testing}
    \vspace{-1.0em}
\end{table}

\paragraph{Results and Discussion} Results 
are presented in Table \label{tab:meta_testing}. We find that SALTED metamorphic testing elicits a number of novel bugs, providing new data points/instances for investigating system errors or comparing system performance.
Unlike other metamorphic tests (e.g., \citet{pit}), the metamorphic testing enabled by the SALTED framework leverages detectors for error checks and is therefore high-precision by default. 

\section{Fixing Salient Long-Tailed Errors}
\label{sec:sec8}

\begin{algorithm}[ht]
 \SetAlgoNoLine
 \SetNoFillComment
 \KwData{Parallel Dataset S of size $n$, token-level Detector A}
 \KwResult{Meta-Corpus M, Templates T of Size $k$}
 \For{i = 1 to n}{
 \tcc{Sentence Pair Selection}
 Apply Detector $A$ on $S_i$; \\
 If $S_i$ has errors: continue; \\
 \tcc{Templatization}
 Else: Templatize $S_i$ and add to $T$ \\ 
 }
\For{i=1 to k}{
 \tcc{Type Substitutions}
 Substitute $T_k$ with Source-Target Token Mappings of the same Type\\
 Store the Generated Sentence Pairs in M}
\caption{Meta-Corpus Generator \label{algo:meta_corpus}}
\end{algorithm}

Data filtering can improve models by removing erroneous examples, but it doesn't guarantee that there are sufficient examples of a \textit{type} to learn correct behavior from. In this section, we explore the use of detectors to generate an example-dense synthetic corpora for fixing model errors via finetuning.

\begin{table*}[ht]
    \begin{tabularx}{\linewidth}{ l X}
        \toprule
    \textbf{Sequence Type} & \textbf{Instance}  \\
        \midrule
Source &  Remind kids to keep their masks up and stay at least \colorbox{yellow}{six feet} apart. \\
Baseline Output &  Erinnern Sie Kinder, um ihre Masken zu halten und bleiben mindestens \colorbox{orange}{sechs Meter} auseinander. \\
Finetuned Output & Erinnern Sie Kinder, um ihre Masken zu halten und bleiben mindestens \colorbox{green}{sechs Fuß} auseinander. \\
        \bottomrule
    \end{tabularx}
    \caption{\textbf{Meta-Corpus Based Finetuning}: Accompanying Table 7, this example shows an case where a token-level error (`feet' $\rightarrow$ `Meter') was fixed (`feet' $\rightarrow$ `Fuß') by applying finetuning using the synthetic Meta-Corpus.}
    \label{tab:finetuning_example}
\end{table*}

\begin{table}
  \label{tab:table3}
  \centering
\setlength\tabcolsep{3.0pt}
  \begin{tabular}{lrrr}
    \toprule
    \textbf{Model} &  \textbf{MC Size} &\textbf{Error Cases} &\textbf{BLEU} \\
    \midrule
    Baseline & None & 19 & 31.4  \\ 
    Finetuned & 10K & 4 & 31.6 \\ 
    Finetuned & 20K & 2 & 31.4 \\  
    Finetuned & 50K & 3 & 31.5 \\  \midrule 
  \end{tabular}
   \vspace{-0.75em}
  \caption{\textbf{Finetuning} on Meta-Corpus leads to reduction in physical units errors. \emph{MC Size} is the size of the Meta-Corpus; BLEU scores are reported on WMT20.}
  \label{tab:meta_finetuning}
   \vspace{-1.0em}
\end{table}

\paragraph{Meta-Corpus} Algorithm \ref{algo:meta_corpus} describes the generation of a synthetic corpus wherein we leverage the detectors to ensure that the generated sentence pairs are correct with respect to a particular measurement. An example illustrating the steps is presented in Table~\ref{tab:meta_corpus}. The algorithm consists of three steps: a sentence-pair selection step where a sentence is selected for templatization if the detector does not deem it erroneous, a templatization step for the tokens in the transformation table within the selected sentence pair and finally generation of new sentence pairs by substituting the templatized tokens with (source, target) tokens of the same type. 


\paragraph{Experiment} We generate a `meta-corpus' (Algorithm \ref{algo:meta_corpus}) using the physical units detector on a random sample of the WMT20 training data of size 1M. We then finetune, for 3 epochs, the best checkpoint of the Standard model using a 1:1 mixture of the sentence pairs sampled from the Meta-Corpus and the general 1M training data, filtered using the same detector. We measure general performance on the WMT20 test set and targeted performance on the translation of physical units on the 100K Mono-Evaluation set. Finetuning learning rates are provided in Appendix \ref{sec:appendix_meta_finetuning}. 


\paragraph{Results and Discussion} The results (Table \ref{tab:meta_finetuning}) show that just 10K or 20K `correct' examples, provided by the Meta-Corpus are sufficient to reduce the number of physical unit errors flagged by the detector, while preserving the general model performance. 
An example of this error fix is presented in Table~\ref{tab:finetuning_example}. 
A limitation of this approach is that it can only fix mistranslations, not dropped content. 
In fact, the few cases that remain in the case of Finetuned model (20K) are the cases where the unit is dropped along with a clause in the source sentence.

\section{Related Work}
\label{sec:sec9}

\paragraph{Quality Estimation for MT} The task of Quality Estimation (QE) is concerned with determining the quality of a translation without access to any reference \cite{speciabook, specia_wmt}. In particular, sentence-level QE allows the development of models which act as metrics in the absence of references (QE-as-a-metric). However, such QE-as-a-metric models still focus on a combined evaluation of adequacy and fluency, rendering them insensitive to the presence of long-tailed errors. E.g., consider the two translations in Table 9. The state-of-the-art COMET \cite{comet} QE-as-a-metric model (detailed in appendix \ref{sec:appendix_comet}) produces a score of \textbf{8.73} for the \textbf{baseline} output and \textbf{6.00} for the \textbf{finetuned} output, even though the latter is clearly the correct translation. In appendix \ref{sec:appendix_comet_insensitivity}, we present further quantitative experiments to illustrate this. Note that SALTED correctly flags these errors.


Further, we claim that this insensitivity to a long-tailed errors is not due to deficient modeling of the particular neural QE model, but due to a fundamental limitation of leveraging neural models such COMET \cite{comet} for evaluation. Even though the recent trend in the NLP community has been towards learning neural metrics, we argue that this paradigm isn't equipped to tackle the problem of salient long-tail evaluation since robust interpolation in neural networks requires many orders of magnitude higher number of parameters than currently employed \cite{isoperimetry}, which implies that evaluation using neural models is likely to remain suspect at the long-tail.

\paragraph{Behavioral Testing for MT} A number of previous works \cite{sit, pit, trans_repair, numerical, ref_testing} have tried to construct tests for eliciting errors in NMT systems' behavior. We present a thorough comparison of SALTED against these works along five dimensions in appendix \ref{sec:appendix_related_work}.

\section{Conclusions}
\label{sec:sec10}

In this paper, we have advocated for and demonstrated the utility of a principled, specifications-based, rule-driven approach to reliably flag salient long-tailed MT errors through high-precision detectors. We introduced an iterative, precision-driven process for developing such detectors and applied it on seven classes of MT errors, eliciting a range of errors from state-of-the-art research and commercial systems. Although the manual development of such detectors incurs significant cost, the resulting payoff is high with the constructed detectors applicable across different systems and datasets. Further, we demonstrated the utility of SALTED framework for four different use cases in MT: for obtaining reliable measurements of salient long-tailed errors in translations of arbitrary monolingual data, for corpus filtering, for system comparisons and for fixing token-level errors through a synthetically generated meta-corpus that teaches the model to learn correct behaviors. We hope that our work serves as a useful step towards more reliable MT.

\bibliography{anthology,custom}
\bibliographystyle{acl_natbib}

\appendix

\section{Designing Detector Algorithms}
\label{sec:appendix_design_detectors}

The development of detectors is a manual, rules-driven iterative process with the goal of constructing a very high precision error detector which could be trusted as a measurement of a specific error category. In section \ref{sec:designing_detectors_main}, we presented an example of this process for a token-level detector (physical units). 
\subsection{False Positive Examples}
\label{sec:appendix_false_positives}

We provide a few examples of the false positives obtained in the first two iterations of constructing the physical units detector, the precision of which is enumerated in Table \ref{tab:iteration}. Examples in Table \ref{tab:detector_errors} show a representative sample of the false positives at each iteration. In the first iteration, the error criteria didn't target numeric measurements only, as a result, we got false positives where the change of unit didn't imply semantic change. In the second iteration, we got errors pertaining to an incomplete transformation table, where `Morgen' wasn't specified as a potential translation for the unit `acres'.

\begin{table*}[ht]
    \begin{tabularx}{\linewidth}{ l X}
        \toprule
    \textbf{Iteration} & \textbf{Source-Translation Instance}  \\
        \midrule
1 &  Closed-circuit cameras watch over \colorbox{yellow}{every inch} of the main street.\\
 &  Closed-Circuit-Kameras wachen über \colorbox{yellow}{jeden Zentimeter} der Hauptstraße. \\ \midrule
1 &  The officiant of the wedding then rushed the family away from the beach, back towards a large house \colorbox{yellow}{several yards} away. \\
 &  Der Beamte der Hochzeit eilte dann die Familie vom Strand weg, zurück in Richtung eines großen Hauses, das \colorbox{yellow}{mehrere Meter} entfernt war. \\ \midrule
 2 &  The city of Anaheim tweeted around 2:30 p.m. that the fire was estimated at \colorbox{yellow}{700 acres}. \\
 &  Die Stadt Anaheim twitterte gegen 14:30 Uhr, dass das Feuer auf \colorbox{yellow}{700 Morgen} geschätzt wurde. \\ \midrule
  2 &  California's footprint was even larger: Fires there have now consumed about \colorbox{yellow}{3.1 million acres} - a modern record. \\
 &  Kaliforniens Fußabdruck war sogar noch größer: Feuer dort haben jetzt etwa \colorbox{yellow}{3,1 Millionen Morgen} verbraucht - ein moderner Rekord. \\
 
        \bottomrule
    \end{tabularx}
    \caption{Examples of \textbf{False Positives} in the first two iterations of the physical units detector.}
    \label{tab:detector_errors}
\end{table*}

\section{Full Suite of Detectors}
\label{sec:appendix_detector_suite}

We provide more details on the implementation of detectors. The process of construction of detectors remains the same as described in section \ref{sec:designing_detectors_main}. For each of the detectors, the iterative process is halted when the precision of the error detector reaches 100. This precision is measured using human evaluation, by randomly sampling 100 error instances obtained by applying the detector on 1M source-translation pairs. The sources for obtaining these translations are obtained by randomly sampling the WMT20 monolingual data.

\subsection{Token-Level Detectors}
\label{sec:appendix_token_detectors}
Token-level detectors rely on the construction of transformation tables or transformation functions, that map source tokens to their potential mappings in the target language. In the next sections, we provide implementation-level details regarding token-level detectors.

\subsubsection{Physical Units Detector}

For physical units detector, the entries in the transformation table contain units associated with distance (miles, meters, centimeter, millimeter, inch, kilometre, feet, yard), area (square kilometre, square metre, acres), weight (kilogram, pound), volume (litres, cubic mm) and temperature (celsius, fahrenheit). A number of derivative units follow automatically: e.g., an error translation of `km/hr' getting translated to `miles/hr' could also be detected using the entry for `km' in the transformation table.

\begin{table}[ht!]
    \begin{tabularx}{\linewidth}{l r}
        \toprule
\textbf{Token Transformation Table Entry} &  \textbf{Type}    \\
        \midrule
dollar $\rightarrow$ dollar, usd, dollars, \textbf{\$}   & text    \\
\textbf{\$} $\rightarrow$ \textbf{\$}, dollar, dollars, usd & sym  \\
 rupees $\rightarrow$ \rupee, rupie, rupien, rupee(s), rs & text  \\
 \rupee $\rightarrow$ \rupee, rupie, rupien, rupee, rupees, rs & sym  \\
        \bottomrule
    \end{tabularx}
    \caption{A partial view of the \textbf{Token Transformation Table} constructed for use in currency detector. Each row comprises of allowed token transformations, along with a token `type' annotation (either symbol or text in this case).}
    \label{tab:cur_transformation_table}
\end{table}

\subsubsection{Currency Detector}

For currency detector, a partial view of the transformation table is presented in \ref{tab:cur_transformation_table}. The full entries in the currency table comprise of 20 currencies. We obtained similar false positives as for the physical units detector in Appendix \ref{sec:appendix_design_detectors} until we didn't allow exceptions for idiomatic expressions (e.g., `pennies on the dollar') or approximations (e.g., `a few dollars').

\subsubsection{Large Numbers} For the large numbers detector, we build a transformation table for the text version of larger numbers (`million(s)`, 'billion(s)', `trillion(s)'). We check for their translations into both text and numeric forms.

\subsubsection{Web Terms} For the detector corresponding to web terms, we make use of both transformation table as well as a transformation function. We check for the correct translation of URLs  (which is copying behavior in this case) extracted from the source as well as the correct translation (again, copying in this case) of web terms such as https, www and ftp. Therefore, for this detector, the transformation table comprises only of identity mappings and the transformation function acting on the extracted URL is also an identity mapping.

\subsubsection{Numerical Values Detector}

The numerical values detector allows transformations of the extracted numerical value into a range of possible translations: time-conversions (e.g., `2:00' to `14:00'), date conversions (e.g., `mm/dd/yyyy' to `dd/mm/yyyy'), separator changes (e.g., `10,000' in English to `10,000' in German) and numeric to text forms (e.g., `12' to `zwölf').    

\subsection{Sequence-Level Detectors}
\label{sec:appendix_sequence_detectors}

\begin{table}[ht!]
    \centering
    \setlength\tabcolsep{4.5pt}
    \begin{tabular}{lrr}
    \toprule 
\textbf{Language} &  \textbf{Coverage} &  \textbf{Hallucinations}   \\
        \midrule
    Russian & 5 & 0 \\
    Dutch & 6  & 1 \\
    Danish & 1 & 6 \\ 
    Swedish & 12 & 0 \\ 
    Spanish & 6 & 0 \\ 
        \bottomrule
    \end{tabular}
    \caption{Number of \textbf{Erroneous Translations} flagged by the language-agnostic sequence-level detectors for translations into multiple languages.}
    \label{tab:multiple_languages}
\end{table}

\subsubsection{Coverage}

As described in section \ref{sec:sec4.2}, for implementing the coverage detector, we make use of alignments obtained through a multilingual BERT-based aligner. To compute the number of unaligned tokens in the source, after computing the alignments we filter the source tokens by removing stop-words compiled from a number of sources as well as by removing punctuation tokens. The coverage detector then flags a translation if the number of unaligned content tokens exceed a threshold. This threshold is bucketized in terms of the source sentence length. We use a threshold of 10 if the input sentence length is less than 50 tokens, 20 if input sentence length is between 50 and 100, 30 if the input sentence length is between 100 and 200 and 40 otherwise.

\begin{table}[ht]
    \begin{tabularx}{\linewidth}{ l}
        \toprule
    \textbf{Source-Translation Instance}  \\
        \midrule

 \colorbox{yellow}{The Cougars are supposed to play No.} \\
  \colorbox{orange}{== Weblinks ==== Einzelnachweise ==} \\ \midrule
 \colorbox{yellow}{Ms. Williams was only seeded No.} \\
  \colorbox{orange}{== Weblinks ==== Einzelnachweise ==} \\  \midrule
 \colorbox{yellow}{"Geomsanaejeon" a.k.a.} \\
  \colorbox{orange}{== Weblinks ==== Einzelnachweise ==} \\ \midrule
 \colorbox{yellow}{Greg Brown ( No.} \\
  \colorbox{orange}{== Weblinks ==== Einzelnachweise ==} \\ \midrule
 \colorbox{yellow}{Downtown L. A.} \\
  \colorbox{orange}{== Weblinks ==== Einzelnachweise ==} \\ 
 
        \bottomrule
    \end{tabularx}
    \caption{Examples of \textbf{Hallucinations} in one of the Commercial Translation Systems (Microsoft). The public API was accessed on January 10, 2021.}
    \label{tab:hallucination_example}
\end{table}

\subsubsection{Hallucinations}

The hallucination detector tries to count the number of oscillatory and natural hallucinations \cite{curious}. The detection of oscillatory hallucinations is done by the following algorithm: if the count of the most frequent bigram in the output exceeds the count of the most frequent bigram in the source by 4 and the count of the most frequent output bigram exceeds 10, then it is flagged as an oscillatory hallucination. For detecting natural hallucinations, we compute the number of the unique sources getting translated to the same output, and the output is deemed as a natural hallucination if 5 or more source sentences, each with different lengths translate to it. The hallucination count is then reported by combining the number of natural and oscillatory hallucinations. Note that both the counts are computed independently of each other. For example, in section \ref{sec:sec5}, we found that one of the commercial translators (Microsoft) incurs 5 natural hallucinations, without incurring any oscillatory hallucinations. For illustration, we present these hallucination cases in Table \ref{tab:hallucination_example}.

\begin{table}[ht!]
    \centering
    \setlength\tabcolsep{8.5pt}
    \begin{tabular}{lr}
    \toprule 
\textbf{Detector} &  \textbf{Error cases}    \\
        \midrule
    Coverage & 444  \\
    Hallucinations & 108 \\
    Physical Units & 133 \\
    Currencies & 22 \\ 
    Large Numbers & 84 \\ 
    Web Content & 30 \\ 
    Numerical Values & 405 \\ \midrule
    Total Errors & 1226 \\
        \bottomrule
    \end{tabular}
    \caption{Number of \textbf{Erroneous Translations} flagged by detectors for the WMT21 News Translation task winning system.}
    \label{tab:wmt21}
\end{table}

\subsection{Test-Phase Evaluation of Detectors}
\label{sec:appendix_test_detectors}

To conduct a test-phase evaluation of detectors (the development iterations are halted when absolute precision is achieved on the large initial development corpus) we vary both the monolingual data as well as the system generating the translations. We translate a separate randomly sampled 250K monolingual corpus using the WMT21 winning system and measure the precision of each of the detectors through human evaluation on the flagged input-output pairs. The precision numbers are presented below for each of the detectors. We obtain absolute precision for each of the detectors on all except one of the detectors: Numerical Values (92.53 percent, with 5 false positives). These false positives from the Numerical Values detector pertained to the handling of fractions in certain non-standard forms, which were parsed incorrectly by the detector (an example is presented in Table \ref{tab:false_positive_numerical_value}).

\begin{table}[ht!]
    \centering
    \setlength\tabcolsep{8.5pt}
    \begin{tabular}{lcc}
    \toprule 
\textbf{Detector} &  \textbf{Error cases}&  \textbf{Precision}    \\
        \midrule
    Coverage & 70 & 100.0 \\
    Hallucinations & 1 & 100.0 \\
    Physical Units & 33 & 100.0 \\
    Currencies & 4 & 100.0 \\ 
    Large Numbers & 9 & 100.0 \\ 
    Web Content & 7 & 100.0 \\ 
    Numerical Values & 67 & 92.53 \\
        \bottomrule
    \end{tabular}
    \caption{Number of \textbf{Erroneous Translations} flagged by detectors for the WMT21 News Translation task winning system on 250K `Test' Monolingual data, alongside Precision as adjudged by human evaluation.}
    \label{tab:test_eval_detectors}
\end{table}

\begin{table*}[ht]
    \begin{tabularx}{\linewidth}{ l X}
        \toprule
    \textbf{Sequence Type} & \textbf{Instance}  \\
        \midrule
Source &  Just as he had lost the first set about 1.1/2 hours earlier but turned things around, with the help of a dip in level from the fourth-seeded Zverev. \\
Translation &  So wie er etwa eineinhalb Stunden zuvor den ersten Satz verloren hatte, aber mit Hilfe eines Levelrückgangs des an vierter Stelle gesetzten Zverev die Wende schaffte. \\\bottomrule
    \end{tabularx}
    \caption{False Positive Example for the Numerical Values Detector}
    \label{tab:false_positive_numerical_value}
\end{table*}

\section{Sequence-Level Detector Applications}
\label{sec:appendix_detector_applications}

We translate the 100K monolingual sentences into 5 different languages using a commercial system (Microsoft) and measure the number of coverage and hallucination errors. We find that the same thresholds used for English-German apply well to the languages in Table \ref{tab:multiple_languages} too, with the flagged outputs exhibiting the related error conditions.

\section{Examples from WMT21 Winning System}
\label{sec:appendix_wmt21_examples}

In this section, we translate the 1M Monolingual Evaluation set using the WMT21 News Translation task winning system. Beam size of 5 was used for generating the translations \footnote{https://github.com/pytorch/fairseq/tree/main/examples/wmt21}. We report the detector error counts in Table \ref{tab:wmt21} and examples for different error categories in Table \ref{tab:wmt21_examples}. Table \ref{tab:wmt21_examples} shows the error instances from different detectors. Here, counts and the examples show that a range of long-tail errors errors persist in the WMT21 system as well, with a \textbf{0.12\% incidence rate}, similar to that of the commercial systems. 

\section{Experimental Details}
\label{sec:appendix_section6_details}

 For experiments, we use fairseq~\cite{fairseq}. Sentencepiece \cite{spm} with a joint token vocabulary of 32K was learned over the training corpus. The Transformer model used, comprising of 6 layers with embedding size 512, FFN layer dimension 4096 and 16 attention heads, was trained for 100 epochs, with the best checkpoint selected using the loss score on the validation set. Additional experimental details are provided in appendix \ref{sec:appendix_section6_details}. For, BLEU, TER, ChrF2++ evaluations SacreBLEU is used \cite{sacrebleu}, for COMET scores the implementation provided by \citet{comet} is used. All models were trained on 8 Nvidia V100 GPUs and a beam size of 5 was used for each evaluation.

\begin{table}[ht!]
    \centering
    \setlength\tabcolsep{8.5pt}
    \begin{tabular}{lr}
    \toprule 
\textbf{Data Source} &  \textbf{Sentence Pairs}    \\
        \midrule
Europarl  & 1,828,521    \\
ParaCrawl &	34,371,306  \\
Common Crawl &	2,399,123 \\
News Commentary	& 361,445 \\
Wiki Titles	& 1,382,625 \\
Tilde Rapid	& 1,631,639 \\
WikiMatrix &	6,227,188 \\ \midrule
Total  & 48,201,847	\\
        \bottomrule
    \end{tabular}
    \caption{The WMT20 \textbf{Data sources} used for training the English-German models in section \ref{sec:sec6}}
    \label{tab:data_source}
\end{table}

\begin{table*}[ht!]
    \begin{tabularx}{\linewidth}{l X}
        \toprule
    \textbf{Detector} & \textbf{Source-Translation Instance}  \\ \midrule
    
Hallucination & Lampard and Mourinho exchanged barbs as old friends became enemiesCredit : PA : Press Association \\
& in Lampard and Mourinho exchanged barbs as old friends became enemiesCredit \colorbox{orange}{: PA : Press Association ( PA : PA : PA : PA : PA : PA : PA : PA ) Lampard} \newline
\colorbox{orange}{and Mourinho exchanged barbs as old friends became enemiesCredit : PA :} \newline
\colorbox{orange}{PA : Press Association ( PA : PA : PA )} \\ \hline

Web Content & Go to the Income Tax Department website by typing \colorbox{yellow}{https://www.incometaxindiaefiling.gov.in/home} in the address bar of your browser. \\
& \colorbox{orange}{incometaxindiaefiling.gov.in/home} in die Adressleiste Ihres Browsers. \\ \hline

Physical Units & Scott McLaughlin came close from a free kick \colorbox{yellow}{30 yards} out. \\
 &  Scott McLaughlin kam mit einem Freistoß aus \colorbox{orange}{30 Metern} in die Enge. \\ \hline  


Large Number & With support from the array of diplomatic associations, A.F.S.A. has so far raised three quarters of a \colorbox{yellow}{million dollars}, Rubin said. The funds are helping defray the legal fees of seven witnesses, covering in full charges that were not paid by the State Department or waived through pro-bono assistance. "It's very moving that current and former Foreign Service officers, most of whom don't have much money, have contributed to help their colleagues," John Bellinger, who served as a legal adviser to the State Department and National Security Council during the George W. Bush Administration, told me. He and a former C.I.A. general counsel, Jeff Smith, represented Ambassador Taylor and Ambassador Mike McKinley. But Taylor, who was not a member of the Foreign Service and was pulled out of retirement to return to Ukraine after Ambassador Marie Yovanovitch was recalled, is not a member of A.F.S.A. - and thus not eligible for its financial aid. He was in Ukraine for only six months. Volker is also not a member of A.F.S.A. And none of the witnesses from the White House, Department of Defense, or the Office of Management and Budget qualifies for its aid, either.\\
& "Es ist sehr bewegend, dass aktuelle und ehemalige Beamte des Auswärtigen Dienstes, von denen die meisten nicht viel Geld haben, dazu beigetragen haben, ihren Kollegen zu helfen", sagte mir John Bellinger, der während der Regierung von George W. Bush als Rechtsberater für das Außenministerium und den Nationalen Sicherheitsrat tätig war. Er und ein ehemaliger C.I.A. General Counsel, Jeff Smith, vertraten Botschafter Taylor und Botschafter Mike McKinley. Aber Taylor, der kein Mitglied des Auswärtigen Dienstes war und aus dem Ruhestand gezogen wurde, um in die Ukraine zurückzukehren, nachdem Botschafterin Marie Yovanovitch zurückgerufen wurde, ist kein Mitglied des A.F.S.A. - und somit nicht für seine finanzielle Unterstützung berechtigt. Er war nur sechs Monate in der Ukraine. Volker ist auch kein Mitglied der A.F.S.A. Und keiner der Zeugen aus dem A.F.S.A., dem Haushalts- und Verteidigungsministerium des Weißen Hauses, qualifiziert\\ \hline  

Coverage & \colorbox{yellow}{James Hill of Diamond Bay loves the directive on his local church message board} : "Thou shalt wear a mask - Hygenesis 20 : 20" \\
 &  thou shalt wear a mask - Hygenesis 20 : 20 "(Du sollst eine Maske tragen - Hygenesis 20 : 20)" \\ \hline

Numerical Value & I lost my husband - this month will be a year on the \colorbox{yellow}{14} - so I'm a single parent. \\
 & Ich habe meinen Mann verloren - diesen Monat wird es ein Jahr sein - also bin ich alleinerziehend. \\ 


\bottomrule
    \end{tabularx}
    \caption{\textbf{Detector Output examples} for the WMT21 Winning System using the 1M WMT20 Monolingual-Evaluation set}
    \label{tab:wmt21_examples}
\end{table*}

\subsection{Data Sources and Filtering}
Table \ref{tab:data_source} lists the data sources used for training the models in section \ref{sec:sec6}. The Monolingual evaluation set was sampled from one of the WMT20 monolingual data sources \footnote{https://data.statmt.org/news-crawl/en/}. Further, the language-id filter for the STD-F baseline in \ref{sec:sec6} was built using the more accurate (larger) version of the fasttext released models \cite{fastext-langid}  \footnote{https://fasttext.cc/docs/en/language-identification.html}.

\subsection{Transformer Training}
\label{sec:appendix_transformer_training}
For each of the models, a dropout of 0.1 was used (including relu-dropout and attention-dropout in \cite{fairseq}). The optimizer used was Adam with the adam-betas parameters set to (0.9, 0.98). Clip-norm of 1.2 was used. For each of the models the encoder-decoder embeddings were tied. Each of the models were trained using a maximum batch size of 4096 tokens. Further, 3K warmup updates were used, with the initial learning rate set to 1e-7 and the learning rate set to 1e-4. The batch size was set to 4096 tokens, and the update frequency was set to 200. In each case, the inverse-sqrt learning rate scheduler was used, along with fp16 mode training.

\subsection{Sentencepiece Vocabulary}
For each of the models, the 32K Unigram LM-based sentencepiece \footnote{https://github.com/google/sentencepiece} vocabulary was constructed by using a character coverage of 0.9995, on 3M randomly sampled sentences from the training corpus.

\subsection{SacreBLEU Configuration signatures}
\label{sec:appendix_sacrebleu_config}

The WMT20 En-De SacreBLEU configuration signature for BLEU computation is nrefs:1|case:mixed|eff:no|tok:13a|smooth:exp| \newline
version:2.0.0, for ChrF2++ the signature is nrefs:1|case:mixed|eff:yes|nc:6|nw:2|space:no| \newline
version:2.0.0 and for TER the signature is nrefs:1|case:lc|tok:tercom|norm:no|punct:yes| \newline asian:no|version:2.0.0.

\subsection{Meta-Corpus Fine-tuning}
\label{sec:appendix_meta_corpus_generation}
For meta corpus generation, the substitutions are made using Algorithm \ref{algo:meta_corpus} only on the (source, target) pairs with one occurrence of the physical unit on each side, i.e. only one occurrence of the physical unit is templatized on each side, as illustrated in table \ref{tab:finetuning_example}.

\subsection{Meta-Corpus Fine-tuning}
\label{sec:appendix_meta_finetuning}

For finetuning, in each case, we use 1K warmup updates, with warmup initial learning rate set to 1e-7 and the learning rate set to 4e-4. Rest of the parameter details remain the same as in Appendix \ref{sec:appendix_transformer_training}.

\subsection{COMET QE-as-a-metric model}
\label{sec:appendix_comet}

The COMET QE-as-a-metric model is built on top of XLM-R (large) \cite{xlmr} and is trained on Direct Assessment (DA) scores from WMT 17-19.

\section{Insensitivity Towards Long-Tailed Errors}
\label{sec:appendix_comet_insensitivity}

We add two experiments to quantitatively substantiate the claim the state-of-the-art QE metrics cannot detect salient long-tailed errors with high precision. 

The \textbf{first experiment} is as follows: We select 100 flagged error cases (from the UN-F baseline in Table 6), of the physical units detector (we further verify that these are indeed salient long-tailed errors). Note that these error cases were obtained by applying the detector on 1M translations. We then obtain the Comet-QE scores for each of those 1M translations. We sort the source-translation pairs based on the Comet-QE scores, and measure how many of the flagged cases are present in the worst-K scoring translations. We tabulate this, i.e. how many of the flagged error cases were present in the worst-K scoring translations in Table \ref{tab:comet_1} below. Even in the 100K worst scoring sentences, only 3 out of 100 erroneous translations were present. This shows that the existing state-of-the-art COMET-QE model is insensitive to the salient long-tailed errors pertaining to physical units. Further, we find that this insensitivity holds true across different salient long-tailed error categories. 

For the \textbf{ second experiment}, instead of sorting the translations based on Comet-QE scores, we measure how many of the erroneous cases were present in the set of translations that scored below a certain threshold. Table \ref{tab:comet_2} presents the results. This shows that the erroneous cases are spread across a range of scores.

\begin{table}
  \centering
\setlength\tabcolsep{1.00pt}
  \begin{tabular}{lc}
    \toprule
    \textbf{K}  & \textbf{Cases in K} \\
    \midrule
    100 & 0  \\
    1000 & 0  \\
    10000 & 0  \\
    100000 & 3 \\ \bottomrule
  \end{tabular}
  \caption{\textbf{Number of flagged Error cases present in the \textbf{K} worst scoring translations, as scored by COMET-QE \cite{comet}}. The erroneous sentences are present very sparsely even in the lowest 100K scoring translations.}
  \label{tab:comet_1}
\end{table}

\begin{table}
  \centering
\setlength\tabcolsep{1.00pt}
  \begin{tabular}{lcc}
    \toprule
    \textbf{Threshold }  &
    \textbf{Sentences } & 
    \textbf{Cases} \\
    \midrule
    0.1 & 251,518 & 26  \\
    0.2 & 346,083 & 36 \\
    0.3 & 522,128 & 70  \\
    0.4 & 649,828 & 79  \\
    0.5 & 742,194 & 88  \\ \bottomrule
  \end{tabular}
  \caption{\textbf{Number of flagged Error cases present in the translations with score less than the threshold, as scored by COMET-QE \cite{comet}}. The erroneous translations are present across a range of scores.}
  \label{tab:comet_2}
\end{table}

\begin{table*}[t]
\centering
\setlength\tabcolsep{4.60pt}
    \begin{tabularx}{\linewidth}{l c c c c c}
        \toprule
    \textbf{Method} & \textbf{Instance-Level} & \textbf{Modularized} & \textbf{Specification-Based} & \textbf{High-Precision} & \textbf{Generative} \\
        \midrule
SIT &  \text{\sffamily x}  & \text{\sffamily x}   & \text{\sffamily x}  & \text{\sffamily x} & \checkmark \\
PatInv &  \text{\sffamily x}  & \text{\sffamily x} & \text{\sffamily x} & \text{\sffamily x}   & \checkmark \\
TransRepair &  \text{\sffamily x}  & \text{\sffamily x} & \text{\sffamily x} & \text{\sffamily x}   & \checkmark \\
RTI &  \text{\sffamily x} & \text{\sffamily x} & \text{\sffamily x} & \text{\sffamily x}   & \text{\sffamily x} \\
SALTED &  \checkmark  & \checkmark   & \checkmark   & \checkmark   & \checkmark \\
        \bottomrule
    \end{tabularx}
    \caption{A comparison of existing Behavioral Testing Methods for NMT along five dimensions. The compared methods are: SIT \cite{sit}, PatInv \cite{pit}, TransRepair \cite{trans_repair} and RTI \cite{ref_testing}.}
    \label{tab:bt_comp}
    \vspace{-1.0em}
\end{table*}

\section{Related Work}
\label{sec:appendix_related_work}

We situate SALTED among previous works in Behavioral testing for NMT along five dimensions in Table \ref{tab:bt_comp}. The five dimensions reflect the operational properties of behavioral testing methods:

\begin{enumerate}
    \item \textbf{Instance-Level}: A method operating at an instance-level requires only the source-translation instance for an error to be adjudged. Typically, Behavioral testing methods rely on input modifications to test the model for errors, thereby requiring the generation of new translations. A method which doesn't work at the instance-level is better suited for exploratory uses than for obtaining targeted error measurements over a given corpus.
    \item \textbf{Specification-Based}: A specification-based method explicitly consumes specifications of correct model behavior. For example, Behavioral testing methods which rely only on consistency measures over translations generated on an input set do not consume an explicit output behavior specification and thereby are hard to translate into actionable measurements.
    \item \textbf{Modularized}: A modular method allows for fine-grained measurements of specific error categories using the same method by separating the concerns of the error detection algorithm and the error type. For example, a method which is not modularized is hard to adapt to a new error type.
    \item \textbf{High-Precision}: A high-precision method produces very few false-positives, ensuring that the generated measurements are trustworthy.
    \item \textbf{Generative}: A generative method allows for the generation of new samples either for metamorphic testing or for data augmentation or error correction.
\end{enumerate}

Table \ref{tab:bt_comp} shows that compared to existing behavioral testing methods, SALTED is more comprehensive, thereby allowing for variety of use cases.

\end{document}